\documentclass[submission,copyright,creativecommons]{eptcs}
\providecommand{\event}{FMAS 2022} 

\usepackage{iftex}

\ifpdf
  \usepackage{underscore}         
  \usepackage[T1]{fontenc}        
\else
  \usepackage{breakurl}           
\fi

\usepackage{tabularx}
\usepackage{listings}
\usepackage{graphicx}
\usepackage{amsmath}
\usepackage{booktabs}

\newcommand{\toolname}{RoTRA}
\newtheorem{definition}{Definition}

\title{Advising Autonomous Cars about the Rules of the Road}
\author{Joe Collenette \qquad Louise A.~Dennis \qquad Michael Fisher
\institute{Department of Computer Science\\University of Manchester, UK}
\email{joe.collenette@manchester.ac.uk} }

\def\titlerunning{Advising Autonomous Cars about the Rules of the Road}
\def\authorrunning{J. Collenette, L. Dennis \& M. Fisher}
\begin{document}
%
\maketitle  
\begin{abstract}
This paper describes (R)ules (o)f (T)he (R)oad (A)dvisor, an agent that provides recommended and possible actions to be generated from a set of human-level rules.  
We describe the architecture and design of RoTRA, both formally and with an example.
Specifically, we use RoTRA to formalise and implement the UK ``Rules of the Road", and describe how this can be incorporated into autonomous cars such that they can reason internally about obeying the rules of the road.
In addition, the possible actions generated are annotated to indicate whether the rules state that the action \emph{must} be taken or that they only recommend that the action \emph{should} be taken, as per the UK Highway Code (Rules of The Road).
The benefits of utilising this system include being able to adapt to different regulations in different jurisdictions; allowing clear traceability from rules to behaviour, and providing an external automated accountability mechanism that can check whether the rules were obeyed in some given situation. 
A simulation of an autonomous car shows, via a concrete example, how trust can be built by putting the autonomous vehicle through a number of scenarios which test the car's ability to obey the rules of the road. 
Autonomous cars that incorporate this system are able to ensure that they are obeying the rules of the road and external (legal or regulatory) bodies can verify that this is the case, without the vehicle or its manufacturer having to expose their source code or make their working transparent, thus allowing greater trust between car companies, jurisdictions, and the general public. 
\end{abstract}

\section{Introduction}
Vehicle manufacturers are aiming to bring fully autonomous cars (AVs) as a product to the wider public.
For fully autonomous cars, trust from both the general public and individual jurisdictions is critical for their successful deployment.
Companies that develop autonomous cars are often unwilling to share the code they have developed and, even when prepared to do so, such code may be opaque to inspection and analysis (if, for instance, it uses deep neural networks to deliver some of the functionality). 
We wish to develop a stronger basis for such trust by enabling autonomous cars to demonstrate that they obey the rules of the road, even in the presence of proprietary or opaque source code.

Jurisdictions throughout the world all have different rules regarding how cars interact with the road, and with other road users.
Having completely separate implemented systems for different regions is inefficient and liable to errors.
We believe that having a modular system, where the rules of the road are kept separate from the other processes the car may be handling, such as visual input processing, will not only allow more efficient designs, but safer, more trustworthy, and more transparent autonomous cars.

Jurisdictions will want to ensure that autonomous cars obey the rules of the road, in a manner similar to human road users.
When autonomous cars are deployed to the wider public, they will be occupying the road with other human drivers.
By obeying the rules of the road, other human drivers will be better able to predict how the autonomous car will behave in a given situation.
The two main issues we tackle are:

\begin{enumerate}
	\item The rules of the road are specific to a jurisdiction;
	\item Certification and assurance requires traceability from regulations to behaviour.
\end{enumerate}
Thus our research question is: \emph{how do we architect a solution that addresses the two problems above but does not require autonomous vehicles to make their source code both available and transparent}?

By addressing these issues we aim to create more trustworthy autonomous cars.
We have developed \toolname{}, which implements a given country's rules of the road and which the autonomous car can query at any point.
An autonomous car can query \toolname{} using its knowledge of the current state of the world to find out what actions (if any) they are  legally required to make, or recommended to make, based on the rules of the road.
\toolname{} makes no assumptions about how the autonomous car interprets the world, so it can be implemented in many different manufacturers' cars. 
From a manufacturer's perspective this provides a modular mechanism that enables their underlying system to receive information about the rules it \emph{should} be obeying at any moment and also to demonstrate adherence to those rules. 
\toolname{} only requires three arguments (context, beliefs, intentions) to be able to generate advice though it is assumed that these arguments are provided within a shared ontology that describes road situations and vehicle actions.
This shared ontology is, in turn, derived from the rules of the road.
Communicating at this level allows reasoning about road rules to occur at the same level of abstraction at which the rules are written.
This has the benefit of making the associated behaviour transparent to stakeholders, and also separating the reasoning from sensor processing, path planning and other lower-level features of the vehicle which enables this reasoning to be evaluated separately.
The development of the tool allows us to address the first issue of ensuring the autonomous vehicle knows the currently applicable rules of the road.

To show the benefits of \toolname{} and how it can be used in conjunction with a certification process, we have developed a number of simple simulated autonomous cars which all consult \toolname{} about the rules of the road. 
We then take some basic scenarios, which require rules to be followed, such as obeying traffic lights. 
The results of the simulations demonstrate both how an autonomous vehicle can flexibly respond to information about road rules (particularly when such rules are recommended rather than required).
At the same time, such a scenario based framework could constitute a ``virtual driving test" --- an autonomous vehicle could be tested to see if it followed the recommendations of \toolname{} appropriately.
The aim of the experiment is to highlight how the second issues of certification and assurance can be tackled using our tool.

\section{Background}

Having an autonomous car obey the rules of the road is not a trivial task, the rules being designed to be read and executed by humans. 



Part of the German rules of the road, Straßenverkehrsordnung (StVO), has been implemented in Linear Temporal Logic (LTL) \cite{rizaldi2017formalising}.
Rizaldi \cite{rizaldi2017formalising} focused on the overtaking sections of the StVO, and their LTL implementation was formally verified using Isabelle/HOL.
We have taken a different approach as we provide the autonomous cars with advice on the actions they can take to bring them in line with the rules of the road.
Additionally we have intentionally kept the autonomous car implementation and rule statements separate.

Similarly Bhuiyan \cite{bhuiyan2020traffic} uses defeasible deontic logic to model the rules of the road.
In this instance they modelled overtaking rules for Australia, specifically the Queensland overtaking traffic rules.
Bhuiyan \cite{bhuiyan2020traffic} evaluated their system by testing whether their rules of the road DDL system could accurately determine if it was safe to overtake.
To determine if it was safe to overtake, the DDL system required information from the world, modelling the environment as an ontology which gathered the necessary knowledge.
To evaluate their model Bhuiyan \cite{bhuiyan2020traffic} simulated a number of overtaking scenarios, and to ensure that the results were truthful they employed domain experts to interpret the simulation. 
Our aim is to simplify the implementation of the rules of the road, and to keep separate the evaluation and knowledge gathering sections.

Enabling autonomous vehicles to conform to traffic law is also not a trivial task. 
Prakken \cite{prakken2017problem} looks at Dutch traffic law and how these rules create challenges in the context of the AI and Law field.
Three possible approaches to tackling this challenge proposed by Prakken \cite{prakken2017problem} are:
\begin{itemize}
\itemsep=0pt
	\item System guarantees compliance;
	\item Using traffic models;
	\item Training to be compliant.
\end{itemize}
The first approach uses model checking to ensure that the autonomous car does not exhibit undesirable properties~\cite{alves2020reliable,alves2021double,Fernandes2017,kamali2017formal,schwammberger2018abstract}. 
The second approach models the traffic, which can include the rules themselves, such as \cite{bhuiyan2020traffic,rizaldi2017formalising}.
The third approach trains a model, via machine learning techniques, to ensure that an autonomous vehicle does not exhibit the undesirable properties \cite{hong2019rules,lee2021deep,shalev2016safe}. 

From these approaches Prakken \cite{prakken2017problem} identifies that the rules of the road need to be formalised and calls for a focus on ``safe and anticipatory'' driving aimed at generating correct behaviour according to the stated rules, as opposed to moral reasoning about ``Trolley problems'' and the like (as typified in work such as Bonnefon's~\cite{bonnefon2016social}).
Our approach aims to allow correct behaviour to be generalised among many manufacturers which can then be validated by external parties.

AI and Law as a area of research has developed approaches to implementing law as a program using a number of techniques \cite{al2016methodology,bench2012history,collenette2020explainable}, such the British Nationality Act \cite{sergot1986british}. 
Among these are rule-based systems, the approach taken here.
Bench-Capon \cite{bench1994defence} makes the case for the use of rule based systems in the AI and Law context and, in particular, the use of rules as tools to help inform a human decision maker without dictating what must happen without question.
We extend this approach to one where the system informs the (potentially) proprietary and (potentially) opaque system within an autonomous vehicle.

Rules governing vehicles are not limited to cars on roads; there is work on obeying the rules that apply to marine vehicles \cite{benjamin2006navigation,enevold2021grounding}, and these approaches have focused on implementing the rules in the decision making process.
Our approach separates the decision making process from the rules and so would be applicable to any autonomous vehicle which needs to know what the rules of the ``road'' are for their given situation.


%

In this paper we assume the autonomous car to be truthful about its current state when communicating with the advice system and that its reasoning component will operate identically with both simulated and real inputs.
When implementing this in the real world this is not something that can always be assumed. 
VW recently had an emissions scandal where the systems deployed in the cars were not truthful to the regulators and specifically where the system changed its behaviour depending upon whether it were in regular use or in a test environment \cite{cavico2016volkswagen}.
Regulation regarding autonomous vehicles is a challenge in its own right \cite{araz2018governing,von2018autonomous} and we do not consider the socio-technical aspects of the problem here, only the technical question of providing a test environment for checking adherence to the rules of the road.

\section{\toolname{}}
We split this section into two parts. In Section~\ref{sec:arch} we describe the architecture underpinning the development of \toolname{}. Then, in Section~\ref{sec:rules}, we show how we have implemented the UK rules of the road using this architecture.
The aim of developing this architecture is to address the first issue we identified of allowing autonomous vehicles to know what the applicable rules of the road are.

\subsection{Overview}
\label{sec:arch}
Given a set of inputs describing the road situation from the point of view of an autonomous vehicle (autonomous vehicle) and its current intended actions, \toolname{}{} returns a set of further actions.

Internally \toolname{} represents the rules of the road as a set of rules which define a set of actions to be performed given the road situation and an autonomous vehicle's intentions.
The road situation is represented as a set of beliefs and a set of intentions (e.g. the autonomous vehicle believes that traffic light ahead is red, and intends to approach a traffic light). 
A context which describes meta-level concepts concerning when the rule should be interpreted is also attached to each rule (e.g., that the given rule only applies in emergency situations).
Actions are labelled with a status, with the meaning of this depending on the jurisdiction (e.g., ``must'' for actions that must be taken and ``should'' for actions that are advised).  

We represent rules in this way based on an analysis of the UK Highway Code.
In this code the majority of rules do not apply in emergency situations, but a few rules exist which do.  
While the context could be represented as an additional belief, it will be clearer to stakeholders wishing to compare the implementation with the written document if context is separated out from those beliefs that describe the current road layout and infrastructure.

The distinction between ``must'' and ``should'' in the recommendations returned to the autonomous vehicle allows flexible reasoning around whether such an action should be taken in a given situation taking into account social norms, the urgency of the autonomous vehicle's current goals, and (un)expected behaviour from other motorists.

We assume a standard first-order logical language, $\mathcal{L}$, built up of constants, variables, terms, and formulae, with the standard connectives and quantifiers: $\land$, $\lor$, $\lnot$, $\implies$, $\forall$, $\exists$.

\begin{definition}[autonomous vehicle Situation] The tuple $\langle Context, \mathit{Beliefs}, Intentions \rangle$ is an \emph{autonomous vehicle situation} where $\mathit{Beliefs}$ and $Intentions$ are sets of ground terms in $\mathcal{L}$, and $Context$ is a constant indicating the meta-level context.

Given a situation, $s = \langle Context, \mathit{Beliefs}, Intentions \rangle $ we define:
\begin{align*}
    context(s) & =  Context \\
    \mathit{beliefs}(s) & =  \mathit{Beliefs} \\
    intentions(s) & =  Intentions
\end{align*}
\end{definition}
An autonomous vehicle communicates its situation to \toolname{} in order to obtain advice about applicable rules of the road.

\begin{definition}[Action Pair] We define the tuple $\langle Label, Action \rangle$ as an \emph{action pair}, consisting of a term in $\mathcal{L}$ (the action -- $Action$) and a constant (the status label -- $Label$). 
\end{definition}

\begin{definition} A \emph{Rule} is a tuple, $\langle Situation, ActionPairs \rangle $
where $Situation$ is an autonomous vehicle situation and $ActionPairs$ is a set of action pairs.

Given a rule, $r = \langle Situation, ActionPairs \rangle $ we define:
\begin{align*}
    context(r) & =  context(Situation) \\
    \mathit{beliefs}(r) & =  \mathit{beliefs}(Situation) \\
    intentions(r) & =  intentions(Situation) \\
    ap(r) & =  ActionPairs
\end{align*}

\end{definition}
The semantics of the terms in $\mathcal{L}$, the context and status labels will depend upon a shared ontology between the autonomous vehicle and \toolname{}.
\begin{definition}[Applicable Rule] 
A rule $r$ is \emph{applicable} in an autonomous vehicle situation $s$ iff:
	$$\begin{array}{c}
		\mathit{beliefs}(r) \subseteq \mathit{beliefs}(s) \land \\ intentions(r) \subseteq intentions(s) \land \\ context(r) = context(s) \end{array}$$
%
%
%
If $r$ is applicable in $s$ then we write, $applicable(s, r)$.\\
Given a set of rules $\mathcal{R}$ and a situation $s$ we define the set of applicable rules from $\mathcal{R}$ in $s$ as:
$$applicable(s, \mathcal{R}) = \{r \in \mathcal{R} \mid applicable(s, r)\}$$
\end{definition}
This allows us to define the set of recommended actions, labelled with their status, given some set of rules and an autonomous vehicle situation.

\begin{definition}Given a set of rules $\mathcal{R}$ and an autonomous vehicle situation $s$ then the set of labelled actions (action pairs) recommended in order to obey the rules is defined as:
$$recommended(s, \mathcal{R}) = \bigcup_{r \in applicable(s, \mathcal{R})} ap(r)$$
\end{definition}
To summarise, we define a number of rules which apply in an associated autonomous vehicle situation consisting of a context, set of beliefs, and set of intentions.
Each rule returns a set of actions labelled with a status when the rule's autonomous vehicle situation is a subset of the situation communicated to it by the autonomous vehicle.
The tool thus allows an agent to find what actions they should do according to the full set of rules defined, based on their current context, set of beliefs, and set of intentions.  At present \toolname{} assumes all terms are ground so simple subset relations suffice for determining applicability.  This assumption was sufficient for our case study.

The actions recommended can be analysed and validated at the rule level, though obviously whether an autonomous vehicle then adheres to the recommended actions will need to be validated separately. 
However, this decomposes the assurance effort into first validating the rules and then validating whether they are followed.

\subsection{Implementation}
\label{sec:rules}

We have implemented our architecture for \toolname{} in Prolog.
Each rule is represented separately as a fact, with 5 variables.

\begin{lstlisting}[basicstyle=\footnotesize\sffamily,breaklines=true]
rule(Name, Context, B, I, A)
\end{lstlisting}
Each rule, when defined, has a name, context, list of beliefs (B), list of intentions (I), and a list of actions, (A).
All applicable rules are found for the given context (L).
Then a list of action pairs is retrieved, sorted, and returned (R).
\begin{lstlisting}[basicstyle=\footnotesize\sffamily,breaklines=true]
getRecommended(C, B, I, R) :-
  findall(X, rule(X,C,_,_,_), L),
  getRecommended(L, C, B, I, [], A),
  sort(A, R).
\end{lstlisting}
The list of action pairs is found by checking each rule individually, and adding the action when the beliefs and intentions of the rule are a subset of the beliefs and intentions passed to the program.
\begin{lstlisting}[basicstyle=\footnotesize\sffamily,breaklines=true]
getRecommended([RName|Tl], C, B, I, A, R) :-
  rule(RName, C, Br, Ir, Ar), ((subset(Br, B), subset(Ir, I)) -> 
             append(Ar, A, Ao); Ao = A),
             getRecommended(Tl, C, B, I, Ao, R), !.
	
getRecommended(RName, C, B, I, A, R) :-
  rule(RName, C, Br, Ir, Ar), ((subset(Br, B), subset(Ir, I)) -> 
             append(Ar, A, Ao); Ao = A),
             getRecommended([], C, B, I, Ao, R), !.
	
getRecommended([], _, _, _, A, Ao) :- Ao = A.	
\end{lstlisting}

\subsubsection{The UK Highway Code}
We have hand analysed all 307 rules, taken from the UK Highway Code \cite{highwayCode}.
During our analysis we note that the UK highway code is not only limited to motor vehicles, but also includes rules for both cyclists and pedestrians.
We determine whether a rule is applicable to autonomous cars before we implement it in a \toolname{} prototype.

An example of an unimplemented rule is rule 37 which specifies that ``Invalid Carriages"\footnote{Typically powered wheelchairs and mobility scooters} must obey signs when on the road, and should be treated as pedestrians when on the pavement.
We have not implemented this rule as \toolname{} is designed with autonomous cars in mind.
Additionally some rules of the road which apply to motorised vehicles will not need to be applied to autonomous vehicles.
Rule 94 specifies that the driver should not wear tinted glasses while driving if there is reduced visibility. 
Autonomous cars are unlikely to tint their vision systems, so there is little point in modelling rule 94 in our system.

If the rules of the road are followed to the letter, there may be some actions which seem unusual to other road users.
Rule 113, allows headlights to be off at night if the road is lit by street lighting.
Driving without your headlights on at night would be concerning to other road users even though this is permitted by the rules of the road.
It is important, therefore, that an autonomous vehicle be allowed to take additional actions, even if not expressly required by the rules.
\toolname{} is not designed to make decisions for the autonomous car, it is designed to inform a decision making process of any design what the current applicable rules of the road are.

During our analysis of the UK Highway Code we have seen that there is an implicit hierarchy to the rules.
Rules that are enforced by humans (e.g., policemen directing traffic in person), override temporary signs which override permanent signs which override everyday rules.
In our implementation we have not modelled this hierarchy explicitly.

The rules of the road also contain a few rules applicable only in emergencies. Correspondingly, the vast majority of rules do not apply in emergency situations.
\toolname{} allows for different rules to apply with the same beliefs and intentions through context.
Each rule is implemented as having a context which applies in a ``standard" driving or an ``emergency" situation.

The UK rules stipulate that actions specified as ``must" have legal backing and those specified as ``should" do not.
We similarly apply ``must" and ``should" to each action.

There are a number of rules that are duplicates of other rules.
Through our analysis of the UK Highway Code, we have noted two more aspects to consider, firstly that an individual rule in the Rules of Road will need multiple implementations to cover everything stated in that rule.
An example of an individual rule needing to split into two Prolog rules is rule 226.
Rule 226 \cite{highwayCode} states
\begin{quote}
	You MUST use headlights when visibility is seriously reduced, generally when you cannot see for more than 100 metres (328 feet).
	You may also use front or rear fog lights but you MUST switch them off when visibility improves (see Rule 236).
\end{quote}
We split this into rule 226a, which states when headlights must be on and fog lights which should be on, and rule 226b which states when fog lights must be off.
Secondly the UK rules of the road are not minimal -- some rules are repeated, with different wordings.
This can also be seen in quote above, which references rule 236.
Rule 236 \cite{highwayCode} states
\begin{quote}
	You MUST NOT use front or rear fog lights unless visibility is seriously reduced (see Rule 226) as they dazzle other road users and can obscure your brake lights. You MUST switch them off when visibility improves.
\end{quote}
In terms of when fog lights must be off, there is no practical difference in the two rules.

We have chosen to represent all rules, including any overlapping rules, which also means that our implementation is not minimal, but is more amenable to checking by stakeholders.

We now give an example of how rules 226 and 236 are converted from human readable form, as shown in the above quotes, to Prolog facts that can be used in the \toolname{} system.
For rule 226 we identify that there are multiple facts that need to be created. 
From there we take the first rule which we name r226a, and identify the necessary beliefs that the car should have in this situation. 
Namely that the visibility is seriously reduced.
Next we identify the actions that need to be suggested to the autonomous vehicle.
For r226a we identify that headlights must be on, and fog lights should be on.
Finely we identify the context in which this rule applies, which is a standard driving context.
We now show the Prolog facts for rule 226.
\begin{lstlisting}[basicstyle=\footnotesize\sffamily,breaklines=true]
	rule(r226a, standard, [visibilitySeriouslyReduced], [], [must-headlights_on,  should-fog_lights_on]).
	rule(r226b, standard, [visibilityClear], [], [must-fog_lights_off]).
\end{lstlisting}

\noindent Our implementation does not make any checks for whether the actions generated by the rules of the road conflict.
The final decision on what to do in any given situation is still on the autonomous car.  Broadly speaking, where two rules conflict then providing only one is legally required (a ``must'' rule) then choice between them should be straightforward.  If two ``should'' rules conflict then the autonomous vehicle has the flexibility to chose which (if either) should be obeyed.  While we don't address the issue of the consistency of the rules of the road here, we note that our formalism would allow consistency checking to take place and the limited number of ``must'' rules would make such an endeavour practical.  
Overall we have implemented 313 rules into our Prolog program.


\subsubsection{Autonomous Vehicle Implementation}
We briefly discuss how \toolname{} might be integrated into an autonomous car.  We make a number of assumptions about the behaviour of the car, namely
we assume that the autonomous car is honest, truthful, and capable of interpreting the actions sent to it.
We make no assumptions about the design or architecture of the autonomous car.
The only requirement is that the autonomous car is capable of running a Prolog instance and has the ability to communicate the beliefs and intentions stated in the rules of the road.
We recognise that some actions that the implementation sends over will be difficult to implement in practice.
For example, in the UK Highway Code, an autonomous car is legally required to be `considerate' to other road users.
Defining what consideration is, and how it can be implemented and measured in a autonomous car is a significantly difficult task, and out of scope of what this paper hopes to achieve.
We also recognise that some beliefs and intentions may be outside the current abilities of autonomous vehicle technology, such as recognising what kind of road the autonomous car is currently driving on, though we argue that these are issues that need to be solved irrespective of the use of \toolname{}.
In Section \ref{sec:Discussion} we discuss how \toolname{} will cope with the real-time requirements of autonomous cars.

We now give a basic example of how \toolname{} works in practice, using an example of an autonomous car that has left its fog lights on after leaving a foggy area.
The autonomous car will send the following query, based on its current beliefs and intentions:

\begin{lstlisting}[basicstyle=\footnotesize\sffamily,breaklines=true]
	getRecommended(standard,  [fog_lights_on, visibility_clear, driving, headlights_on], [], Actions).
\end{lstlisting}
\toolname{} then returns the following list back to the autonomous car:
\begin{lstlisting}[basicstyle=\footnotesize\sffamily,breaklines=true]
	Actions = [must-consideration_others, must-drive_care_attention, must-fog_lights_off, must-not_drive_dangerously].
\end{lstlisting}
The autonomous car now has information which it can incorporate into its decision making process.
The rules of the road state that the autonomous car is legally required to turn off its fog lights, but the decision to do so (or not) is still the choice of the autonomous vehicle.

\section{Experiment}
We have created a basic simulation framework to test \toolname{} against simple autonomous cars.
The aim of this set of experiments is to highlight the benefits of autonomous cars implementing \toolname{}, via their behaviour in a number of scenarios.
By creating these experiments we highlight how the second issue we identified surrounding certification and assurance can be mitigated through an autonomous car which implements \toolname{}.

\begin{figure}
	\centering
	\includegraphics[width=0.8\linewidth]{"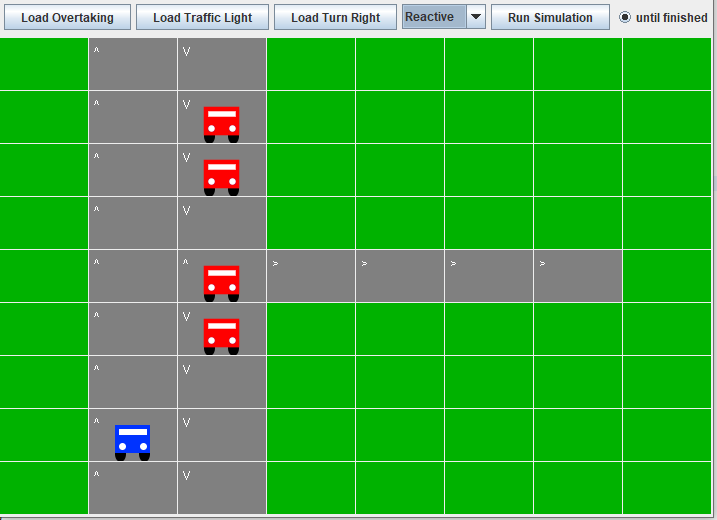"}
	\caption{Simulator in JAVA showing turn right scenario, with the blue car representing the agent to test, red cars are generic traffic. \label{fig:Simulator}}
\end{figure}

The simulation is a grid world, where each cell in the world contains information about its contents, such as whether the cell can be driven on.
Each agent in the simulation is able to observe a small portion of the surrounding world.
Outside of the observable world, the agent is not given any information.
All agents decide what their actions will be simultaneously, and these are then executed simultaneously, constituting a simulation step.
The worlds are infinite, with roads looping back and the north connecting to the south, and east to west.
Figure~\ref{fig:Simulator} provides a visual representation of the simulator running a scenario, with 5 agents (cars) running.

The situations are designed to test the ability of the autonomous car to obey a selection of the rules of the road.
The group of scenarios can be thought of as a (basic) autonomous car driving test.

In our experimentation, we use 3 different situations to test different aspects of the rules of the road.
These are:
\begin{description}
	\item[Traffic Light] A simple road with traffic light on it, where the autonomous car is required to pass by the traffic light. 
	
	This scenario tests whether the autonomous car obeys traffic lights.
	\item[Overtake] A two lane road, where the autonomous car is approaching a slow moving vehicle and can choose to overtake the vehicle in front.

	This scenario tests whether the autonomous car can overtake safely, and then return to the left lane.
	\item[Right Turn] A two way road, one lane in each direction, with a turn off half way up, shown in Figure~\ref{fig:Simulator}.
	The autonomous car intends to turn right and needs to do so safely, avoiding traffic coming in the opposite direction.
	
	This scenario tests the autonomous car's ability to handle right turns within traffic.
\end{description}

\noindent The autonomous cars that we have developed aim to complete the test in the shortest time possible, the agents that drive the car implement \toolname{} to inform their decision making process.
We vary the extent to which the autonomous vehicle decision making process pays attention to the actions that are recommended to it by \toolname{}.
The agents that will be driving the cars are:

\begin{description}
	\item[Reactive] Passes its current state to \toolname{}, and obeys all recommended actions that are returned.
	\item[Morally bankrupt] Passes the current state to \toolname{}, and only obeys actions that have the ``must" prefix. 
	This autonomous car intends to only obey the bare legal requirements of the rules of the road. Otherwise it intends to complete the task as fast as possible.
\end{description}
After running these agents in the 3 different situations, we will observe their behaviour and note the behaviours we see, and how well the agent obeys the rules of the road.
Observation summaries for each agents behaviour is shown in Table~\ref{tab:Results}.
\begin{table}[t]
	\centering
	\caption{Observed behaviours of the 2 different agents after each experiment has been conducted\label{tab:Results}}
	\begin{tabularx}{\textwidth}{rXX}
		\toprule
		Experiment & Reactive Agent & Morally bankrupt agent\\ \midrule
		Traffic Light & Obeys all traffic lights, and continues along road  & Obeys all traffic lights, and continues along road\\
		Overtake & Overtakes the slower car in a safe manner, and returns into the left lane when safe. & Overtakes the slower car safely, however does not return to left lane remains in overtaking lane\\
		Turn Right & Waits for  traffic to pass, then makes a safe right turn & Turns in front of traffic, making a unsocial right turn\\ \bottomrule
	\end{tabularx}
\end{table}
Firstly in the overtake experiment we observed that the morally bankrupt agent does not return to the left lane after overtaking a slow moving vehicle.
Not moving into the left lane after an overtaking violates the rules of the road, but that action is only a ``should'' so there is no legal obligation to do so.  
Secondly we observed that in the Turn Right scenario that the Morally Bankrupt agent leaves no room for traffic coming the other way when turning right.
If we were in a position to say whether the agents passed a virtual driving test, then based on the observed behaviours we would pass the reactive agent.
The Morally Bankrupt agent would not pass the test, again based on the observed behaviours of the agent because while it is, strictly speaking. driving legally it is violating the more abstract concepts of considerate and safe driving -- we discuss this further in section~\ref{sec:Discussion}.
We can also see what the agents believed to be true in the world, and so can judge  whether the violation is because the agent has an incorrect belief about the state of the world or if the rules have been intentionally ignored.
The evidence generated from these tests can then be passed to certifiers, and the developers of the agent so that they can improve their autonomous cars ability to follow the rules of the road.

In addition we tested a third agent ``Proactive" which was the same as the ``Reactive" agent, but in addition to finding the recommended actions for the current state, Proactive also found the recommended actions for the next predicted state.
We found no observable difference between the Proactive and Reactive agents in any scenario we tested.
The UK rules of the road are often described in the form of ``intending to achieve'' followed by the actions to achieve the particular state.
For example within the rules regarding traffic lights, one rule states that if you intend to a approach traffic light and it is red then you must stop at the white line. Having to supply intentions as part of such rules already forces agents to look ahead to guide future actions, thus explaining why the Proactive agent provided no advantage over the Reactive one.


In summary we have implemented \toolname{} into a number of agents, which drive a simulated autonomous vehicle in a number of different scenarios.
From these simple scenarios we can identify where rules of the road have been broken, and provide evidence where an agent breaks them.



\section{Discussion \label{sec:Discussion}}
\subsection{Abstract Actions}
As already noted some of the actions recommended by \toolname{} such as driving with consideration of others and driving with care and attention are not straightforward to interpret.  However, as our experiments showed following more specific ``should'' recommendations can act as a proxy for these more abstract concepts.  Our Morally Bankrupt car failed its ``driving test'' because it failed to leave room for oncoming traffic when turning right hence violated both these directives.  It is outside the scope of this paper to consider deeply how these more abstract actions should be implemented but we observe that following ``should'' recommendations in many cases may form a part of this.

\subsection{Scalability}
Each individual autonomous car will have a copy of the program that is local to that car.
Meaning that the number of autonomous cars does not affect the speed which the program operates.
When \toolname{} is used in a real time situation the autonomous car will need to provide the state of the world consistently in given time intervals.
This means that the implementation of \toolname{} needs to be fast and efficient.
We explored the time taken for our Prolog implementation to return actions.
The code was run on a iMac 2020 (Big Sur 11.6.4), with a 3.3Ghz Intel Core i5 with 8GB of ram.
The example code is taken from the experiment where the autonomous car is approaching a red light.

\begin{lstlisting}[basicstyle=\footnotesize\sffamily,breaklines=true]
getRecommended(standard,[vehicleSafe,headlightsOff,allChildrenUsingChildSeatAsRequired,canReadNumberPlate,exitClear,dualCarriageWay,vehicleDoesntFitsInCentralReservation,roadAheadClear,fuel,driving,completeOvertakeBeforeSolidWhiteLine,routePlanned,lightRed,allPassengersWearingSeatBeltsAsRequired,sidelightsOff],[approachingTrafficLight],Actions)).
\end{lstlisting}

The output of the Prolog $time$ command was:
\begin{lstlisting}[basicstyle=\footnotesize\sffamily,breaklines=true]
	% 2,453 inferences, 0.000 CPU in 0.001 seconds (77% CPU, 6086849 Lips)
\end{lstlisting}

\noindent Our implementation of \toolname{} in Prolog runs quickly, which in turn would allow autonomous cars in a real time situation to poll the program with small time intervals.

\subsection{Verifying Vehicle Control}
While we do not carry out any verification here,  formal verification can be used in conjunction with \toolname{}.
 %
Since the component controlling an autonomous vehicle, for example an agent, has access to \toolname{} a straightforward property to formally prove is that the agent never ignores `must' directives.
We can formally verify this of an agent controlling the vehicle by using a tool such as AJPF~\cite{MCAPLjournal} and, in doing so, we pass some of the burden of safe decision making on to the rules of the road.
Assuming our autonomous vehicle is guaranteed to \emph{always} follow directives from \toolname{} then, as long as the representation of the rules is correct, the vehicle is guaranteed to follow the rules of the road.
Similarly, we might formally verify that our agent controlling the vehicle does not ignore `should' directives, unless it has a good reason for doing so.

\subsubsection{Monitoring Vehicle Behaviour}
Runtime verification is a mechanism for assessing a formal property concerning the system's behaviour \emph{as it is running}. 
If we embed \toolname{} into such a runtime monitor, for example using standard robotic techniques~\cite{ferrando2020rosmonitoring}, then the monitor can watch the vehicle behaviour and recognise when the vehicle chooses some behaviour at odds with the directives provided by \toolname{}.
Once such behaviour is recognised, the monitor might either record this, report it to an external body, or use this to provide suggestions and guidance to the vehicle control.

\subsubsection{Guidance}
Especially if a \emph{human} driver is controlling the vehicle then \toolname{} can provide guidance on what \emph{must} or \emph{should} be done. 
Using similar mechanisms to the runtime monitors above we can recognise the situation and, through simple explanations (e.g., as ~\cite{koeman2019why} supplies for cogntive agent programs), provide guidance to a human driver.
The driver is, of course, free to make their own decision but a clear and short explanation of the rule and directive that applies can potentially have a beneficial effect.
\subsection{Relationship to Ethical Reasoning}

The past decade has witnessed an explosion of interest in enabling autonomous and AI systems to reason ethically (See~\cite{nallur20,tolmeijer2021survey} for recent surveys of the field).  Proposed frameworks, architectures and implementations include those based on symbolic reasoning (e.g.,~\cite{bringsjord2006toward}), those based on machine learning (e.g.,~\cite{BalakrishnanBouneffoufMatteiRossi2019}) and hybrid approaches (e.g.,~\cite{ecoffet2021reinforcement}).  On an architectural level the ethical reasoning can be deeply embedded in the system, having a say in the generation of actions (e.g.,~\cite{anderson2018value}) or it can exist as a separate \emph{governor module} that may veto or modify actions generated via some other method (e.g.,~\cite{arkin2008governing}).

A useful question to ask is whether legal reasoning is just a special case of ethical reasoning that should be embedded within an appropriate ethical reasoning framework, rather than existing separately.  An important observation here is that this may depend upon the system requirements -- does obeying the law always take primacy over other ethical considerations?  In our case, except in a very few situations, it is clear from the framing of the UK Highway Code that nearly all rules may be set aside if it is ethically imperative to do so.  As such, it is not unreasonable to suppose that the legal rules should be considered as part of the ethical reasoning (it is ethical to obey the law) but may be set aside if other ethical considerations are more important.  An example of a system of this kind is laid out in~\cite{dennis2016formal} which considers the case of an Unmanned Aircraft which may chose to ignore the Rules of the Air as laid out by the UK Civil Aviation Authority, if doing so will save lives. The work in~\cite{millan-blanquel2020} adapts this system to AVs.  Although this latter work does not explicitly include legal concerns in its implementation, it does consider the relationship of ethical reasoning to legal reasoning in the autonomous vehicle context as part of its wider discussion.

\toolname{}, as a system that provides \emph{information} about what the law advises, rather than \emph{deciding} upon a legal action, was conceived as a component part of a reasoning system which decides upon action using information from multiple sources.  There is therefore no reason why the output from \toolname{} could not be used by an ethical reasoning process operating within the autonomous vehicle.  Indeed, there are good reasons why an autonomous vehicle should employ some form of ethical reasoning and it would be logical for the output of \toolname{} to be used by that  process.

\section{Conclusion}
We have developed a framework, \toolname{}, which allows a system to retrieve recommendations on which actions to take based on the current state of the world.
\toolname{} has been defined formally, and in Prolog.
The aim of developing \toolname{} was to address the first issue identified of allowing autonomous vehicles to query what the current rules of the road are.
To implement \toolname{} in practice we have analysed all 307 UK rules of the road.
We then implemented the rules in Prolog so that can be used within the \toolname{} framework.
We have highlighted how rules are converted from human readable form, to their Prolog fact implementations, noting how some rules need to split into multiple facts, and some rules are not applicable to autonomous cars.

To show how \toolname{} can work in practice, we created a set of experiments, which implemented \toolname{} in a simulation environment involving a basic autonomous car.
The experiments showed some of the benefits of the system, in addressing the second issue identified of certification and assurance.
These benefits include the ability to see where autonomous cars have broken the rules of the road, and the ability to provide evidence of these rules being broken.
Other benefits of \toolname{} are that rules from multiple jurisdictions can be changed on the fly.
Users of the tool will be able to record what state of the agent believes itself to be in, without exposing the source code of the agent.

There are a number of avenues to build on this work. 
The issue of trust has not been solved by this paper, but moved from the autonomous car to the implemented tool and the certification process.
We are well placed to build the trust in the implemented tool, by formally verifying the implementation we have created. 
We acknowledge that the certification process will need to lead by and implemented by governments, who would create the laws needed to ensure that all cars go through the process. 
The UK government does recognise that laws and processes regarding autonomous cars need to be developed, which was made explicit by recognising that the rules of the road need to change to allow autonomous vehicles \cite{gov2021AV}.

One of the assumptions that we have used in this paper is that the information the autonomous receives is reliable. 
There are number of reasons why this might not be the case;
Input sensors may construct incorrect beliefs of the world to car.
Automotive manufacturers may conceal or distort the true state of the world to the program.
\toolname{} will need further experimentation in these scenarios to find the most effective way of addressing the truthfulness assumption.

In this paper we have also discussed issues such the scalability of the tool and the relationship the program has to ethical reasoning.
Further discussion was given about how verifiability will provide a rich avenue of exploring the effectiveness of the tool, along with how \toolname{} can be used in other scenarios such as supporting human drivers.

To summarise we have developed a tool, \toolname{}, which allows vehicles to know what actions they should take for a given situation.
The separation of the tool from the decision making process of the agent allows third parties to test whether the agent is obeying the rules of the road, without the need to expose the source code of the agent.

\paragraph{Acknowledgements:} This work is supported by EPSRC, through  EP/V026801 (\emph{Verifiability Node}) and EP/W01081X (\emph{Computational Agent Responsibility}), and by the Royal Academy of Engineering, under their \emph{Chairs in Emerging Technologies} scheme.

For the purpose of open access, the author has applied a Creative Commons Attribution (CC BY) licence (where permitted by UKRI, ‘Open Government Licence’ or ‘Creative Commons Attribution No-derivatives (CC BY-ND) licence may be stated instead) to any Author Accepted Manuscript version arising.

All data for the following paper can be found at the following two github respositories:
\begin{itemize}
  \item \url{https://github.com/JoeCol/RulesOfTheRoadExperiments}
  \item \url{https://github.com/JoeCol/SimpleCarSimulator}
\end{itemize}

\bibliographystyle{eptcs}
\bibliography{rules}

\end{document}